# A COMPARATIVE STUDY OF IN-AIR TRAJECTORIES AT SHORT AND LONG DISTANCES IN ONLINE HANDWRITING

Carlos Alonso-Martinez (1), Marcos Faundez-Zanuy (1), Jiri Mekyska (2)
(1) ESUP Tecnocampus (Pompeu Fabra University) Av. Ernest Lluch 32, 08302 Mataró, Spain.
(2) Department of Telecommunications, Faculty of Electrical Engineering and Communication, Brno University of Technology, Technicka 10, 616 00 Brno, Czech Republic
calonso@tecnocampus.cat , faundez@tecnocampus.cat, mekyska@feec.vutbr.cz

**ABSTRACT**

*Introduction* Existing literature about online handwriting analysis to support pathology diagnosis has taken advantage of in-air trajectories. A similar situation occurred in biometric security applications where the goal is to identify or verify an individual using his signature or handwriting. These studies do not consider the distance of the pen tip to the writing surface. This is due to the fact that current acquisition devices do not provide height formation. However, it is quite straightforward to differentiate movements at two different heights: a) short distance: height lower or equal to 1 cm above a surface of digitizer, the digitizer provides x and y coordinates. b) long distance: height exceeding 1 cm, the only information available is a time stamp that indicates the time that a specific stroke has spent at long distance. Although short distance has been used in several papers, long distances have been ignored and will be investigated in this paper.

*Methods* In this paper, we will analyze a large set of databases (BIOSECURID, EMOTHAW, PaHaW, Oxygen-Therapy and SALT), which contain a total amount of 663 users and 17951 files. We have specifically studied: a) the percentage of time spent on-surface, in-air at short distance, and in-air at long distance for different user profiles (pathological and healthy users) and different tasks; b) The potential use of these signals to improve classification rates.

*Results and conclusions* Our experimental results reveal that long-distance movements represent a very small portion of the total execution time (0.5 % in the case of signatures and 10.4% for uppercase words of BIOSECUR-ID, which is the largest database). In addition, significant differences have been found in the comparison of pathological versus control group for letter l in PaHaW database (p=0.0157) and crossed pentagons in SALT database (p=0.0122)

## 1. INTRODUCTION

Speech and handwriting are probably the most difficult tasks performed by human beings, because they differentiate us from animals. Handwriting requires very fine motor skills, probably more so than speech, because some animals can imitate human sounds but no animal can write. In addition, we learn to speak first and then we learn how to read and write, when the brain is more mature.

Handwriting analysis is a good way to study the human brain in a non-invasive way. This knowledge, once acquired, can be applied to artificial systems that emulate the human brain. We consider that handwriting movements are more complex by far than what has been analyzed in the past. In fact some parts of the movements have been neglected. With this paper we will analyze this kind of movements, which will be defined in posterior sections as in-air at long distance. This kind of movements can be used to improve artificial intelligence for biometric applications such as health and security (Faundez-Zanuy et al. 2013), (Lopez-de-Ipina et al. 2015), (Sesa-Nogueras et al. 2016), (Rosenblum and Luria, 2016).

In the past, the analysis of handwriting had to be performed in an offline manner. Only the writing itself (strokes on a piece of paper) were available for analysis. Nowadays, modern capturing devices like digitizing tablets and pens or online whiteboards, can gather data without losing its temporal dimension. When spatiotemporal information is available, its analysis is referred to as online. A typical modern digitizing tablet (Figure 1) not only gathers the x-y coordinates that describe the movement of the writing device as it changes its position, but it can also collect other data, mainly the pressure exerted by the writing device on the writing surface; the azimuth (the angle of the pen in the horizontal plane) and the altitude (the angle of the pen with respect to the vertical axis), see (Figure 2). From now own, x-y coordinates, pressure, azimuth and altitude will be referred to as *features of the handwriting*.

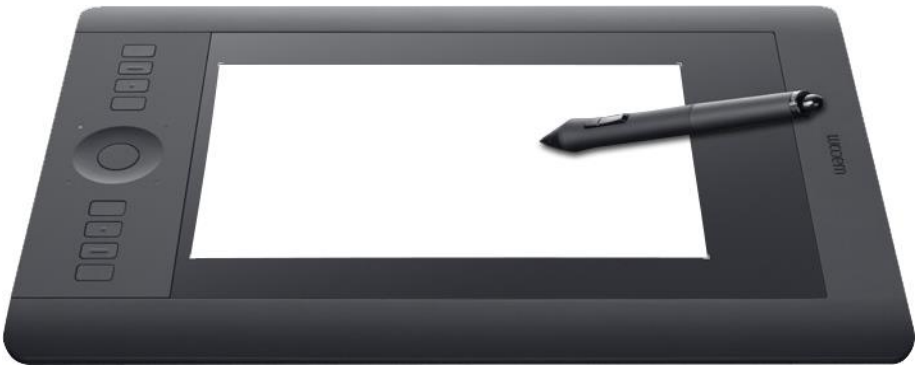

**Figure 1. Intuos Pro L digitizing tablet and pen.**

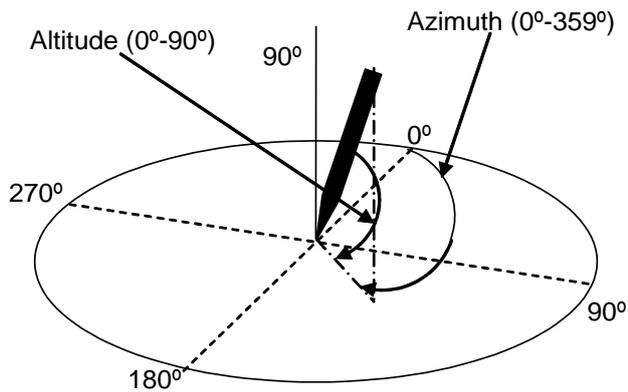

**Figure. 2 Azimuth and altitude angles of the pen with respect to the plane of the writing surface**

A very interesting aspect of the modern online analysis of handwriting is that it can consider information gathered when the writing device was not exerting pressure on the writing surface. Thus, the movements performed by the hand while writing a text can be split into two classes:
(a) On-surface trajectories (pen-downs), corresponding to the movements executed while the writing device is touching the writing surface. Each of these trajectories produces a visible stroke. We will call this kind of movement on-surface.
(b) In-air trajectories (pen-ups), corresponding to the movements performed by the hand while transitioning from one stroke to the next one. During these movements, the writing device exerts no pressure on the surface. This class can be split into two subsets:
  a. In-air at short distances (in-air$_S$), when the distance from the tip of the pen to the writing surface is lower or equal to 1 cm. In this case, the digitizing device can track the (x, y) coordinates during the pen movement.
  b. In-air at long distances (in-air$_L$), when distances from the tip of the pen to the writing surface are higher than 1 cm. In this case, the digitizing device is not able to track the movements and we only know the time spent at high distance.

In our previous research, we have focused on on-surface and in-air$_S$ movements discarding in-air$_L$ movements because they do not provide the same amount of data as the previous ones. In fact, the unique parameters are just the number of strokes at long distance and time spent at long distance. For instance, in (Sesa et al. 2012) we applied information theory to demonstrate that on-surface and in-air$_S$ contain almost the same amount of information and they are not redundant. This was an important milestone because in-air trajectories had received almost no attention at all, even in online approaches where spatiotemporal information is available.

Figure 3 shows two examples of on-surface and in-air$_S$ trajectories taken from two executions of the pentagon test performed by two different writers from the Emothaw database.

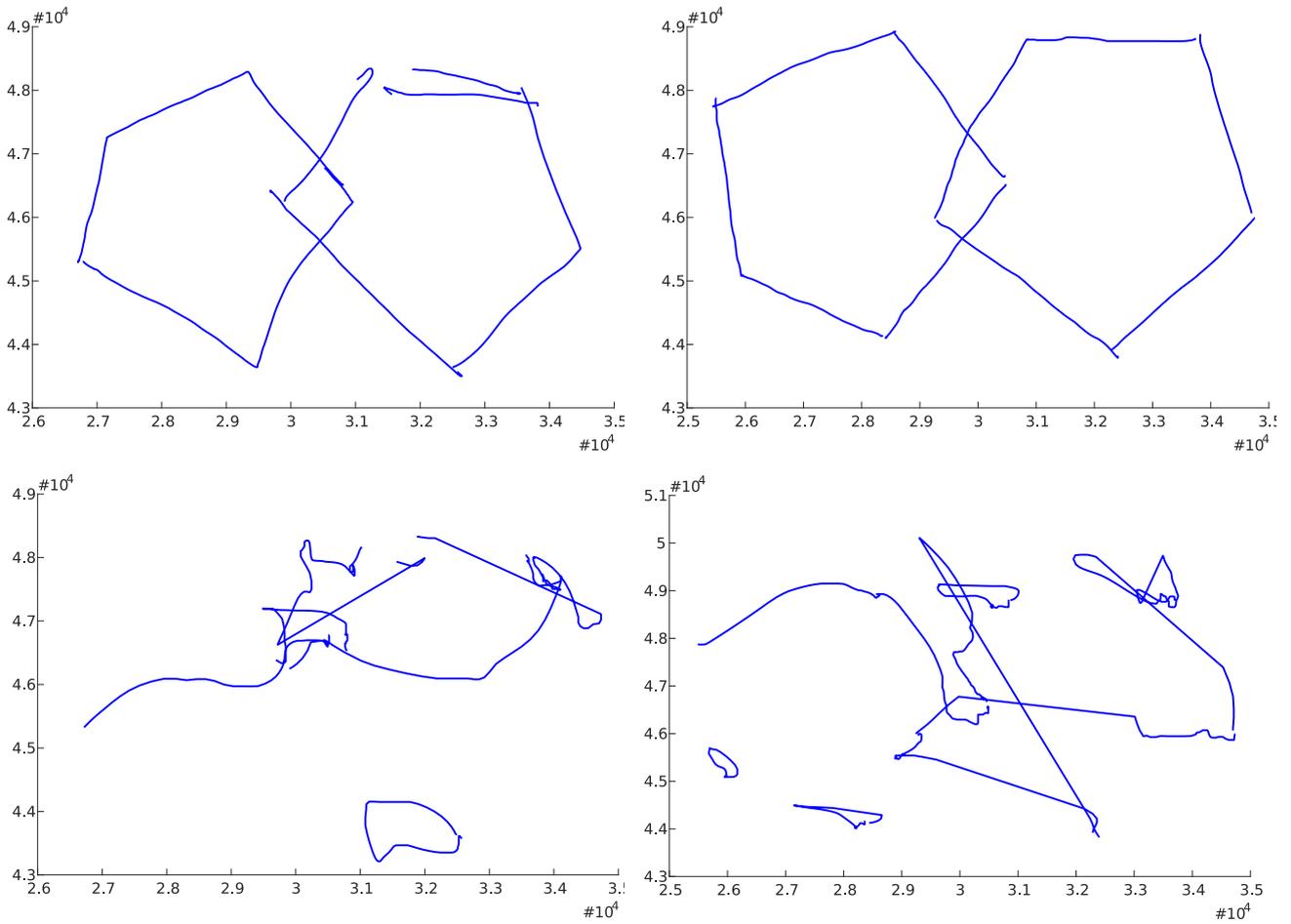

**Figure 3. On-surface (top) and in-air (bottom) trajectories from two executions of TWO CROSSED PENTAGONS**

In-air$_L$ can be detected looking at the time-stamp provided by the digitizing tablet. During in-air$_L$ time the tablet is unable to track the tip of the pen and no samples are acquired. Nevertheless, time stamp is increasing and the next time that the pen touches the surface, the samples are stored again in the file and the time jump can be detected. Figure 4 shows the difference of consecutive time stamps for an example file. For most of the samples (on-surface and in-air$_S$) this value is small (typically two units). However, there are some peaks, which correspond to in-air$_L$ movements. Figure 4 reveals 11 strokes of the type in-air$_L$. Sometimes this time is abnormally long. This is probably due to some acquisition problem, where the user started to speak with the database acquisition supervisor for minutes. We will label these cases and will not include them in the average computation of time spent at in-air$_L$. We consider these cases when time in-air$_L$ is greater than 70% of the total time. In particular, we have found this phenomenon in 5 files from analyzed databases (total amount of analyzed files is 17951 files).

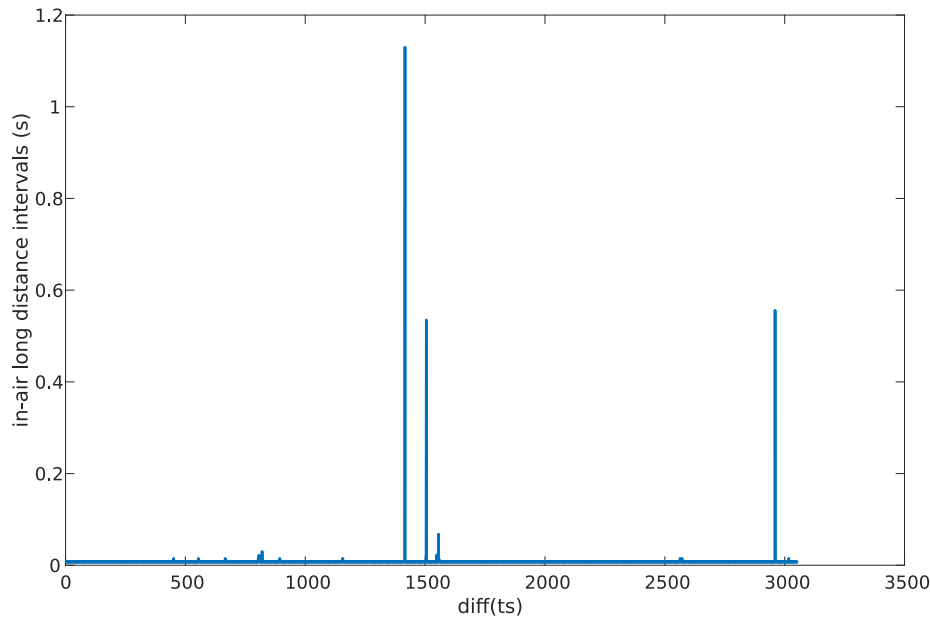

**Figure 4. Time stamp difference of consecutive samples for an example of accepted file from PaHaW database task write *lektorka* word twice**.

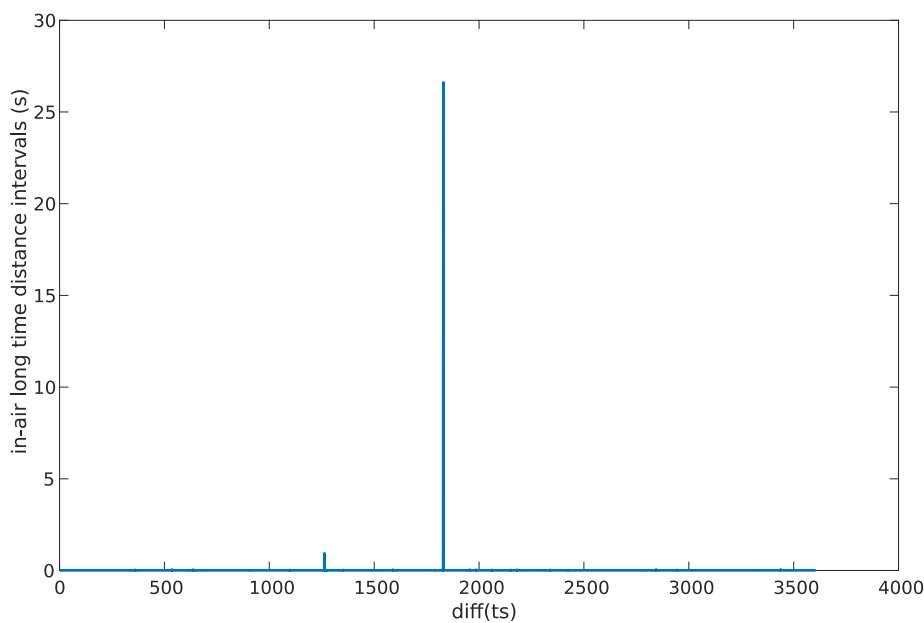

**Figure 5. Time stamp difference of consecutive samples for an example of discarded file from PaHaW database task write *lektorka* word twice.**

## 2. EXPERIMENTAL DATABASES
In this paper, we have analyzed a set of different databases that contain different tasks and user profiles. The databases share the existence of handwritten tasks. In this section, we will summarize the main characteristics of the analyzed databases.

### 2.1 BIOSECUR-ID
This database is a multimodal biometric one and includes eight biometric traits: speech, iris, face (still images and videos), handwritten signature and handwritten text, fingerprints, hand and keystroking. This database acquired inside the BiosecurID project, was developed by a consortium of six Spanish Universities, more details can be found in (Fierrez et al. 2010). With respect to handwriting and signatures, this database defines five different tasks: a Spanish text in lowercase, ten digits written separately, sixteen Spanish words in upper-case, four genuine signatures and one forgery of the three precedent subjects.

## 2.2 EMOTHAW

As described in (Likforman-Sulem et al. 2015), this database includes samples of 129 participants who are classified on the basis of their emotional states: anxiety, depression, and stress or health, this classification is assessed by the Depression–Anxiety–Stress Scales (DASS) questionnaire. Seven tasks are recorded through a digitizing tablet: pentagons and house drawing, words in capital letters copied in handprint, circles with left and right hand, clock drawing, and one sentence copied in cursive writing.

## 2.3 PAHAW

The Parkinson's Disease Handwriting Database (PaHaW) consists of multiple handwriting samples from 37 Parkinson's Disease patients and 38 gender and age matched controls. Eight different tasks were recorded through a digitizing tablet: spiral drawing, letters, words, and a sentence. The details about this database can be found in (Drotár et al. 2016).

## 2.4 OXIGEN-THERAPY

This database described in (Fiz et al. 2015) includes eight patients with hypoxemia who performed two tasks: house and clock drawing, before and after breathing 30 minutes with $O_2$ with the aim of evaluating changes in psychomotor functions.

## 2.5 SALT

As described in (Garre el alt. 2016) the database includes samples of 52 participants: 23 with Alzheimer's Disease, 12 with mild cognitive impairment (MCI) and 17 healthy controls. Seven tasks were recorded: crossed pentagons, spiral, 3D house, clock drawings and spontaneous, copied and dictated handwriting.

## 3. EXPERIMENTAL RESULTS

The first experiments consisted of analyzing the three kinds of time in absolute and relative values as well as the number of strokes in all the scenarios. Tables 1 to 5 summarize the results for the analyzed databases. It is worth remarking that different databases contain different tasks described in the previous section.

For a given user, the number of strokes is an integer number. However, the table shows the average number of strokes for a specific database and task (in addition to the number of strokes done by the whole set of users split by the number of users). This number is not integer any more

Table 1. BIOSECUR-ID database. Time in absolute units and relative time in parenthesis.

|  | Time | | | Strokes | | |
|---|---|---|---|---|---|---|
| Task | On-surface | In-air$_S$ | In-air$_L$ | On-surface | In-air$_S$ | In-air$_L$ |
| genuine signature | 2857.6 (79.6%) | 715.4 (19.9%) | 17.5 (0.5%) | 6.62 | 5.94 | 0.32 |
| skilled forgeries | 5447.9 (68.5%) | 2373.4 (29.9%) | 128.5 (1.6%) | 6.58 | 6.21 | 0.63 |
| lower case words | 110445.1 (55.9%) | 76454 (38.7%) | 10644.4 (5.4%) | 335.01 | 367.16 | 33.16 |
| numbers | 3677.3 (53.6%) | 3071.1 (44.7%) | 117.0 (1.7%) | 11.66 | 11.46 | 0.79 |
| uppercase words | 73608.8 (54.4%) | 47756.2 (35.3%) | 14073.4 (10.4%) | 313.49 | 343.29 | 30.81 |

Experimental results of BIOSECUR-ID database, which is the largest one according to the number of users and files, reveal that in-air$_L$ is almost negligible in the case of signatures, but interestingly, it is three times larger for skilled forgeries than for genuine signatures. For uppercase words the time in-air$_L$ is larger than for the other tasks but still quite modest (10.4%). Thus, this kind of movement is less important than the other two and can probably be ignored without sacrificing a lot of information. For the other databases, a statistical test will be performed after presenting the experimental results.

Table 2-a. EMOTHAW database (depression). Time in absolute units and relative time in parenthesis.

|  | Time | | | Strokes | | |
|---|---|---|---|---|---|---|
| Task | On-surface | In-air$_S$ | In-air$_L$ | On-surface | In-air$_S$ | In-air$_L$ |
| two-pentagon | 11394.0 (55.5%) | 7755.8 (37.7%) | 1393.3 (6.8%) | 9.26 | 13.15 | 9.47 |
| house | 18765.4 (53.6%) | 13933.1 (39.8%) | 2329.4 (6.7%) | 23.74 | 33.00 | 20.97 |
| capital letters | 15789.4 (51.0%) | 13112.3 (42.4%) | 2050.1 (6.6%) | 59.79 | 65.91 | 12.15 |
| loops with left hand | 10183.9 (97.7%) | 215.8 (2.1%) | 21.3 (0.2%) | 1.26 | 0.41 | 0.21 |
| loops with right hand | 8542.7 (98.9%) | 58.6 (0.7%) | 39.3 (0.4%) | 1.18 | 0.21 | 0.06 |
| clock | 14228.8 (45.0%) | 14905.2 (47.2%) | 2468.7 (7.8%) | 27.35 | 36.91 | 21.44 |
| sentence | 15288.8 (60.4%) | 8052.4 (31.8%) | 1958.5 (7.8%) | 41.24 | 47.41 | 11.09 |

Table 2-b. EMOTHAW database (stress). Time in absolute units and relative time in parenthesis.

| | Time | | | Strokes | | |
|---|---|---|---|---|---|---|
| Task | On-surface | In-air$_S$ | In-air$_L$ | On-surface | In-air$_S$ | In-air$_L$ |
| two-pentagon | 11283.0 (55.0%) | 7768.6 (37.9%) | 1444.1 (7.1%) | 9.41 | 13.89 | 11.39 |
| house | 18868.4 (52.5%) | 14378.6 (40.0%) | 2685.6 (7.5%) | 25.45 | 35.14 | 21.32 |
| capital letters | 15732.3 (50.1%) | 13555.7 (43.1%) | 2135.2 (6.8%) | 60.80 | 67.09 | 12.04 |
| loops with left hand | 10648.5 (97.3%) | 233.5 (2.1%) | 66.6 (0.6%) | 1.59 | 0.77 | 0.95 |
| loops with right hand | 9264.1 (99.3%) | 40.0 (0.4%) | 23.9 (0.3%) | 1.13 | 0.14 | 0.04 |
| clock | 14481.5 (44.8%) | 14934.1 (46.2%) | 2896.2 (9.0%) | 27.63 | 37.80 | 21.41 |
| sentence | 15756.6 (59.4%) | 8539.8 (32.2%) | 2215.8 (8.4%) | 42.55 | 48.95 | 10.84 |

Table 2-c. EMOTHAW database (anxiety). Time in absolute units and relative time in parenthesis.

| | Time | | | Strokes | | |
|---|---|---|---|---|---|---|
| Task | On-surface | In-air$_S$ | In-air$_L$ | On-surface | In-air$_S$ | In-air$_L$ |
| two-pentagon | 11474.7 (57.3%) | 7135.2 (35.6%) | 1420.7 (7.1%) | 8.70 | 12.75 | 10.16 |
| house | 18871.9 (53.6%) | 13683.5 (38.8%) | 2672.7 (7.6%) | 23.77 | 32.89 | 18.95 |
| capital letters | 16010.0 (50.9%) | 13356.9 (42.5%) | 2082.7 (6.6%) | 60.48 | 66.39 | 10.96 |
| loops with left hand | 10248.4 (96.9%) | 224.3 (2.1%) | 103.0 (1.0%) | 1.57 | 0.79 | 0.96 |
| loops with right hand | 8793.2 (99.3%) | 35.6 (0.4%) | 23.9 (0.3%) | 1.11 | 0.13 | 0.04 |
| clock | 14175.3 (46.5%) | 13487.9 (44.3%) | 2811.3 (9.2%) | 26.27 | 35.48 | 19.54 |
| sentence | 15676.5 (59.9%) | 8402.2 (32.1%) | 2107.5 (8.0%) | 41.96 | 48.14 | 10.38 |

Table 2-d. EMOTHAW database (control). Time in absolute units and relative time in parenthesis.

| | Time | | | Strokes | | |
|---|---|---|---|---|---|---|
| Task | On-surface | In-air$_S$ | In-air$_L$ | On-surface | In-air$_S$ | In-air$_L$ |
| two-pentagon | 10256.0 (49.7%) | 8670.5 (42.1%) | 1684.9 (8.2%) | 10.13 | 15.27 | 12.91 |
| house | 17468.1 (49.0%) | 15150.2 (42.5%) | 3044.5 (8.5%) | 26.63 | 36.23 | 22.27 |
| capital letters | 15699.2 (48.9%) | 13721.2 (42.8%) | 2677.1 (8.3%) | 61.46 | 67.84 | 11.68 |
| loops with left hand | 9737.1 (98.5%) | 133.3 (1.3%) | 17.7 (0.2%) | 1.18 | 0.30 | 0.41 |
| loops with right hand | 8992.1 (98.4%) | 123.2 (1.3%) | 23.9 (0.3%) | 1.07 | 0.09 | 0.04 |
| clock | 12365.6 (38.9%) | 16180.8 (50.9%) | 3229.5 (10.2%) | 27.13 | 37.25 | 22.63 |
| sentence | 15660.0 (53.6%) | 9539.6 (32.6%) | 4024.3 (13.8%) | 42.41 | 49.43 | 11.43 |

Table 3-a. PAHAW database (control). Time in absolute units and relative time in parenthesis.

| | Time | | | Strokes | | |
|---|---|---|---|---|---|---|
| Task | On-surface | In-air$_S$ | In-air$_L$ | On-surface | In-air$_S$ | In-air$_L$ |
| spiral | 18665.8 (98.6%) | 171.5 (0.9%) | 103.2 (0.5%) | 1.40 | 1.97 | 1.94 |
| letter l | 8077.8 (57.6%) | 3868.3 (27.6%) | 2069.6 (14.8%) | 5.21 | 18.16 | 15.50 |
| bigram le | 10545.9 (71.2%) | 2998.4 (20.2%) | 1274.3 (8.6%) | 5.13 | 14.03 | 11.00 |
| word les | 12309.1 (69.2%) | 3513.0 (19.7%) | 1977.7 (11.1%) | 5.29 | 15.11 | 11.82 |
| word lektorka | 14931.2 (73.0%) | 3238.1 (15.9%) | 2279.8 (11.1%) | 7.00 | 16.97 | 12.00 |
| word porovnat | 13071.5 (74.4%) | 3356.7 (19.1%) | 1139.4 (6.5%) | 8.08 | 18.08 | 11.82 |
| word nepopadnout | 8757.5 (83.8%) | 1512.5 (14.5%) | 179.0 (1.7%) | 5.29 | 8.47 | 4.50 |
| sentence | 14481.3 (58.4%) | 7457.9 (30.1%) | 2844.6 (11.5%) | 15.24 | 31.87 | 19.34 |

Table 3-b. PAHAW database (Parkinson patients). Time in absolute units and relative time in parenthesis.

| | Time | | | Strokes | | |
|---|---|---|---|---|---|---|
| Task | On-surface | In-air$_S$ | In-air$_L$ | On-surface | In-air$_S$ | In-air$_L$ |
| spiral | 24057.4 (95.4%) | 618.3 (2.4%) | 536.6 (2.2%) | 2.03 | 6.78 | 7.31 |
| letter l | 8928.1 (63.8%) | 4132.5 (29.5%) | 939.1 (6.7%) | 5.51 | 16.08 | 12.59 |
| bigram le | 12143.2 (69.1%) | 4094.1 (23.3%) | 1330.4 (7.6%) | 5.57 | 17.08 | 13.76 |
| word les | 14702.7 (69.6%) | 4093.1 (19.4%) | 2330.9 (11.0%) | 5.76 | 19.22 | 15.54 |
| word lektorka | 17716.2 (76.3%) | 36045.0 (15.5%) | 1890.1 (8.2%) | 7.22 | 17.97 | 12.49 |
| word porovnat | 14690.6 (75.8%) | 3808.9 (19.6%) | 891.1 (4.6%) | 8.86 | 18.11 | 11.00 |
| word nepopadnout | 9784.0 (79.8%) | 2115.7 (17.2%) | 365.6 (3.0%) | 6.76 | 11.30 | 5.86 |
| sentence | 16176.5 (58.2%) | 8252.3 (29.9%) | 3300.1 (11.9%) | 16.57 | 36.81 | 23.62 |

Table 4-a. OXIGEN-THERAPY database (before O$_2$). Time in absolute units and relative time in parenthesis.

| | Time | | | Strokes | | |
|---|---|---|---|---|---|---|
| Task | On-surface | In-air$_S$ | In-air$_L$ | On-surface | In-air$_S$ | In-air$_L$ |
| house | 32699.0 (49.6%) | 22184.8 (33.7%) | 11033.8 (16.7%) | 28.88 | 131.13 | 141.29 |
| clock | 20144.0 (40.2%) | 21824.0 (43.6%) | 8104.3 (16.2%) | 27.25 | 94.13 | 79.00 |

Table 4-b. OXIGEN-THERAPY database (after O$_2$). Time in absolute units and relative time in parenthesis.

| | Time | | | Strokes | | |
|---|---|---|---|---|---|---|
| Task | On-surface | In-air$_S$ | In-air$_L$ | On-surface | In-air$_S$ | In-air$_L$ |
| house | 26572.1 (53.6%) | 18429.1 (37.1%) | 4606.5 (9.3%) | 27.70 | 74.57 | 51.96 |
| clock | 16619.8 (46.4%) | 16007.8 (44.7%) | 3197.6 (8.9%) | 25.21 | 57.21 | 37.29 |

Table 5-a. SALT database (DCLI). Time in absolute units and relative time in parenthesis.

| | Time | | | Strokes | | |
|---|---|---|---|---|---|---|
| Task | On-surface | In-air$_S$ | In-air$_L$ | On-surface | In-air$_S$ | In-air$_L$ |
| crossed pentagons | 18292.8 (60.2%) | 8497.3 (27.9%) | 3612.6 (11.9%) | 10.00 | 20.33 | 27.50 |
| spiral | 8219.3 (99.0%) | 26.75 (0.3%) | 60.8 (0.7%) | 1.42 | 1.75 | 2.25 |
| 3D house | 33503.83 (52.0%) | 19388.6 (30.1%) | 11534.3 (17.9%) | 29.50 | 49.17 | 50.42 |
| clock | 18931.9 (31.2%) | 24807.2 (40.9%) | 16917.0 (27.9%) | 26.67 | 52.17 | 70.50 |
| spontaneous sentence | 16500.3 (48.8%) | 14322.9 (42.4%) | 2966.5 (8.8%) | 40.67 | 47.75 | 15.33 |
| sentence copied | 26535.4 (49.3%) | 21918.3 (40.7%) | 5404.9 (10.0%) | 57.58 | 69.08 | 29.58 |
| sentence dictation | 20710.7 (59.1%) | 11717.8 (33.4%) | 2633.0 (7.5%) | 43.25 | 50.08 | 16.33 |

Table 5-b. SALT database (Alzheimer). Time in absolute units and relative time in parenthesis.

| | Time | | | Strokes | | |
|---|---|---|---|---|---|---|
| Task | On-surface | In-air$_S$ | In-air$_L$ | On-surface | In-air$_S$ | In-air$_L$ |
| crossed pentagons | 21535.4 (48.4%) | 15430.1 (34.6%) | 7555.4 (17.0%) | 14.05 | 28.00 | 48.14 |
| spiral | 11312.2 (88.7%) | 1108.8 (8.7%) | 327.2 (2.6%) | 1.71 | 1.67 | 2.52 |
| 3D house | 40341.6 (47.3%) | 30465.8 (35.8%) | 14386.2 (16.9%) | 31.55 | 55.23 | 75.77 |
| clock | 24524.7 (36.1%) | 33060.4 (48.6%) | 10420.8 (15.3%) | 29.36 | 48.41 | 50.95 |
| spontaneous sentence | 19555.9 (48.6%) | 17090.1 (42.4%) | 3606.1 (9.0%) | 37.23 | 44.05 | 17.09 |
| sentence copied | 34023.8 (45.1%) | 33451.3 (44.4%) | 7951.2 (10.5%) | 54.32 | 69.50 | 35.95 |
| sentence dictation | 26640.6 (52.7%) | 20723.6 (41.0%) | 3189.6 (6.3%) | 44.27 | 54.86 | 20.45 |

Table 5-c. SALT database (control). Time in absolute units and relative time in parenthesis.

| | Time | | | Strokes | | |
|---|---|---|---|---|---|---|
| Task | On-surface | In-air$_S$ | In-air$_L$ | On-surface | In-air$_S$ | In-air$_L$ |
| crossed pentagons | 17077.7 (50.1%) | 13085.8 (38.4%) | 3918.6 (11.5%) | 11.88 | 36.47 | 36.65 |
| spiral | 6198.3 (91.0%) | 426.6 (5.4%) | 251.3 (3.6%) | 1.63 | 3.06 | 2.94 |
| 3D house | 29170.5 (43.3%) | 26094.5 (38.7%) | 12152.4 (18.0%) | 30.82 | 72.24 | 68.12 |
| clock | 18986.1 (30.2%) | 31299.1 (49.8%) | 12547.1 (20.0%) | 29.94 | 71.38 | 71.06 |
| spontaneous sentence | 14990.5 (43.8%) | 14648.8 (42.8%) | 4566.4 (13.4%) | 35.41 | 56.88 | 31.12 |
| sentence copied | 24684.2 (45.5%) | 23968.8 (44.1%) | 5654.8 (10.4%) | 53.59 | 78.00 | 37.53 |
| sentence dictation | 19531.1 (56.9%) | 13131.1 (38.2%) | 1676.5 (4.9%) | 38.71 | 50.24 | 16.76 |

From all the databases related to diseases, we computed the Mann-Whitney U test between study and control groups to determine the existence of statistically significant difference ($p<0.05$) in the studied features (time and strokes). The results are shown in the following tables.

Table 6-a. Emothaw (Depression/Control). $T_S$= time on-surface, $T_{AS}$=time In-air$_S$, $T_{AL}$=time in-air$_L$, Strokes$_S$=strokes on-surface, Strokes$_{AS}$=strokes on-air$_s$, Strokes$_{AL}$=strokes on-air$_L$.

| Task | p $T_S$ | p $T_{AS}$ | p $T_{AL}$ | p Strokes$_S$ | p Strokes$_{AS}$ | p Strokes$_{AL}$ |
|---|---|---|---|---|---|---|
| two-pentagon | 0.4316 | 0.3082 | 0.0589 | 0.1374 | 0.0731 | 0.0561 |
| house | 0.7329 | 0.0495 | 0.5002 | 0.0315 | 0.0774 | 0.4217 |
| capital letters | 0.5771 | 0.5771 | 0.8744 | 0.0904 | 0.2317 | 0.5994 |
| loops with left hand | 0.7613 | 0.2380 | 0.1292 | 0.2954 | 0.2742 | 0.1542 |
| loops with right hand | 0.6592 | 0.5316 | 0.7322 | 0.5067 | 0.5005 | 0.7322 |
| clock | 0.1267 | 0.6293 | 0.2196 | 0.6641 | 0.7739 | 0.9688 |
| sentence | 0.8992 | 0.3273 | 0.1849 | 0.3794 | 0.2870 | 0.5764 |

Table 6-b. Emothaw (Anxiety/Control). ). $T_S$= time on-surface, $T_{AS}$=time In-air$_S$, $T_{AL}$=time in-air$_L$, Strokes$_S$=strokes on-surface, Strokes$_{AS}$=strokes on-air$_s$, Strokes$_{AL}$=strokes on-air$_L$.

| Task | p $T_S$ | p $T_{AS}$ | p $T_{AL}$ | p Strokes$_S$ | p Strokes$_{AS}$ | p Strokes$_{AL}$ |
|---|---|---|---|---|---|---|
| two-pentagon | 0.2429 | 0.1020 | 0.1678 | 0.1546 | 0.1010 | 0.1505 |
| house | 0.4564 | 0.0417 | 0.4086 | 0.0652 | 0.1777 | 0.2550 |
| capital letters | 0.3770 | 0.6374 | 0.3503 | 0.1731 | 0.1888 | 0.7751 |
| loops with left hand | 0.7711 | 0.1575 | 0.1723 | 0.1560 | 0.1429 | 0.2017 |
| loops with right hand | 0.9374 | 1 | 1 | 0.9822 | 0.9762 | 1 |
| clock | 0.0414 | 0.1540 | 0.2410 | 0.4462 | 0.5801 | 0.8294 |
| sentence | 0.7234 | 0.4259 | 0.1296 | 0.5392 | 0.3971 | 0.2913 |

Table 6-c. Emothaw (Stress/Control). ). $T_S$= time on-surface, $T_{AS}$=time In-air$_S$, $T_{AL}$=time in-air$_L$, Strokes$_S$=strokes on-surface, Strokes$_{AS}$=strokes on-air$_s$, Strokes$_{AL}$=strokes on-air$_L$.

| Task | p $T_S$ | p $T_{AS}$ | p $T_{AL}$ | p Strokes$_S$ | p Strokes$_{AS}$ | p Strokes$_{AL}$ |
|---|---|---|---|---|---|---|
| two-pentagon | 0.5665 | 0.4886 | 0.3429 | 0.6173 | 0.4131 | 0.3678 |
| house | 0.5221 | 0.2565 | 0.4705 | 0.5562 | 0.7621 | 0.9188 |
| capital letters | 0.4741 | 0.9907 | 0.7934 | 0.2769 | 0.4367 | 0.4662 |
| loops with left hand | 0.3859 | 0.1625 | 0.2801 | 0.1466 | 0.1498 | 0.3173 |
| loops with right hand | 0.4795 | 0.7184 | 1 | 0.6875 | 0.6875 | 1 |
| clock | 0.0241 | 0.7401 | 0.4670 | 0.6199 | 0.6496 | 0.6623 |
| sentence | 0.6819 | 0.5753 | 0.1011 | 0.7034 | 0.5335 | 0.4764 |

We can observe in table 6-a (Depression/Control) how in crossed pentagon task, the values are very close to the threshold for long distance time and strokes. In house draw, the near time and on-surface strokes show statistical significance. In table 6-b (Anxiety/Control), house draw shows again that near distance time is statistically signficant. Finally, in table 6-c (Stress/Control) we obtain p<0.05 for on-surface time in clock draw only.

Table 7. PaHaW Wilconxon test (Parkinson/Control). ). $T_S$= time on-surface, $T_{AS}$=time In-air$_S$, $T_{AL}$=time in-air$_L$, Strokes$_S$=strokes on-surface, Strokes$_{AS}$=strokes on-air$_s$, Strokes$_{AL}$=strokes on-air$_L$.

| Task | p $T_S$ | p $T_{AS}$ | p $T_{AL}$ | p Strokes$_S$ | p Strokes$_{AS}$ | p Strokes$_{AL}$ |
|---|---|---|---|---|---|---|
| Spiral | 0.3947 | 0.5621 | 0.0939 | 0.2857 | 0.0919 | 0.0949 |
| letter l | 0.4614 | 0.5529 | 0.0157 | 0.2390 | 0.3611 | 0.2718 |
| bigram le | 0.3015 | 0.0403 | 0.5671 | 0.0090 | 0.1173 | 0.1710 |
| word les | 0.3015 | 0.3166 | 0.6601 | 0.2941 | 0.2453 | 0.4385 |
| word lektorka | 0.5166 | 0.9440 | 0.3019 | 0.8111 | 0.6928 | 0.4744 |
| word porovnat | 0.3878 | 0.7226 | 0.4025 | 0.3778 | 0.9239 | 0.7963 |
| word nepopadnout | 0.5780 | 0.1776 | 0.2836 | 0.0630 | 0.1287 | 0.2538 |
| sentence | 0.2000 | 0.5850 | 0.9612 | 0.3229 | 0.2720 | 0.5773 |

As is shown in table 7, for PaHaW database we obtain statistically significant results in letter l long-distance time and in bigram le for near-distance time and on-surface strokes.

Table 8. OXYGEN THERAPY (pre/post $O_2$). ). $T_S$= time on-surface, $T_{AS}$=time In-air$_S$, $T_{AL}$=time in-air$_L$, Strokes$_S$=strokes on-surface, Strokes$_{AS}$=strokes on-air$_s$, Strokes$_{AL}$=strokes on-air$_L$.

| Task | p $T_S$ | p $T_{AS}$ | p $T_{AL}$ | p Strokes$_S$ | p Strokes$_{AS}$ | p Strokes$_{AL}$ |
|---|---|---|---|---|---|---|
| House | 0.8968 | 0.8764 | 0.9174 | 0.9174 | 0.8968 | 0.8968 |
| Clock | 0.9218 | 0.8936 | 0.9077 | 0.8665 | 0.8795 | 0.8795 |

For this database the times and number of strokes don't show statistical significance and do not seem to offer a valid classification pattern between pre and post $O_2$ results.

Table 9-a. SALT (Alzheimer/Control). ). $T_S$= time on-surface, $T_{AS}$=time In-air$_S$, $T_{AL}$=time in-air$_L$, Strokes$_S$=strokes on-surface, Strokes$_{AS}$=strokes on-air$_s$, Strokes$_{AL}$=strokes on-air$_L$.

| Task | p $T_S$ | p $T_{AS}$ | p $T_{AL}$ | p Strokes$_S$ | p Strokes$_{AS}$ | p Strokes$_{AL}$ |
|---|---|---|---|---|---|---|
| crossed pentagons | 0.0303 | 0.1609 | 0.0122 | 0.3941 | 0.6604 | 0.0891 |
| spiral | 0.0063 | 0.5132 | 0.1995 | 0.9185 | 0.1338 | 0.1869 |
| 3D house | 0.0677 | 0.1370 | 0.1297 | 0.3493 | 0.7533 | 0.0720 |
| clock | 0.1071 | 0.1984 | 0.1785 | 0.5526 | 0.2033 | 0.6256 |
| spontaneous sentence | 0.1878 | 0.0524 | 0.9210 | 0.3875 | 0.8316 | 0.8761 |
| sentence copied | 0.0096 | 0.1080 | 0.1096 | 0.5612 | 0.3954 | 0.2629 |
| sentence dictation | 0.0132 | 0.0721 | 0.0920 | 0.2510 | 0.3953 | 0.1604 |

Table 9-b- SALT (MCI/Control). ). $T_S$= time on-surface, $T_{AS}$=time In-air$_S$, $T_{AL}$=time in-air$_L$, Strokes$_S$=strokes on-surface, Strokes$_{AS}$=strokes on-air$_s$, Strokes$_{AL}$=strokes on-air$_L$.

| Task | p $T_S$ | p $T_{AS}$ | p $T_{AL}$ | p Strokes$_S$ | p Strokes$_{AS}$ | p Strokes$_{AL}$ |
|---|---|---|---|---|---|---|
| crossed pentagons | 0.1915 | 0.4925 | 0.1915 | 0.2758 | 0.1688 | 0.5643 |
| spiral | 0.0968 | 0.5358 | 0.0865 | 0.4290 | 0.0889 | 0.0973 |
| 3D house | 0.2069 | 0.5500 | 0.3879 | 0.6729 | 0.7734 | 0.5206 |
| clock | 0.4437 | 0.9445 | 0.0738 | 1 | 0.8892 | 0.2854 |
| spontaneous sentence | 0.5500 | 0.8421 | 0.9119 | 0.4124 | 0.5496 | 0.6094 |
| sentence copied | 0.1501 | 0.4384 | 0.7735 | 0.2777 | 0.6260 | 0.8075 |
| sentence dictation | 0.3640 | 0.7398 | 0.2878 | 0.3407 | 0.4784 | 0.5203 |

In table 9-a (Alzheimer/Control), we can observe how on crossed-pentagons draw, statistical significance can be found in on-surface time and long-distance time. Also, on-surface time presents significance on the sentence copied. No results with p<0.05 were obtained for mild cognitive impairment (MCI)/Control (table 9-b).

4. **DISCUSSION**

Although most of the results in previous tables are not significant, even for on-surface and in-airS information, we should point out that this kind of measurements offers a large set of features that can be extracted, such as speed and acceleration of trajectories, complexity measurements, etc., extracted from coordinates x,y. In fact, a classifier would not be based on a single measurement. It will take advantage of a set of measurements. Thus, high p values for on-surface and in-airS do not imply the impossibility to perform a classification. These values are provided just for comparison purpose with in-airL values. In-airL extracted features are limited to time and number of strokes. Thus, the analysis of relevance of this information is simpler.

Nevertheless this paper points out the tasks and pathologies where more potential improvements can be achieved, because in some tasks $p<0.05$ has been obtained.

Looking at the experimental results of pathologies, we can establish that in-airL movements are not relevant but there are some exceptions: crossed pentagons task for depression patients in Emothaw, which is near significance ($p=0.0589$ for time and $p=0.0561$ for strokes), letter l task for PaHaW database ($p=0.0157$ for time) and crossed pentagons task for Alzheimer/Control comparison ($p=0.0122$ for time). We consider these results especially interesting because crossed pentagons are a very useful measurement in pathological analysis, in fact, it is the only drawing that subjects must perform in the well-established mini-mental state examination, also known as the Folstein test (Folstein et al., 1975).

## 5. CONCLUSIONS

One of the main goals of this paper was to study if in-air$_L$ information can be discarded in handwritten tasks analysis. Looking at the experimental results we can conclude that little time is spent by healthy writers at long distance so most of the information is contained on-surface and in-air$_S$ distances. This implies that the development of a new acquisition device able to track x and y coordinates and long distances will probably not be very useful, because few samples will be acquired in this condition. However, experimental results reveal that time spent at long distance is more than three times higher for skilled forgeries than for genuine signatures. This opens a possible research line in security biometrics. A similar consideration can be established for the number of strokes, which is doubled in the case of skilled forgeries with respect to short distance in-air movements. Thus, the existence of long distance movements can be indicative of a signature forgery.

On the other hand, when looking at pathologies we have found statistically significant differences in the pentagons tasks for Alzheimer/control comparison. This result opens the possibility of investigating in-air at long distance movements further.

## 6. ACKNOWLEDGEMENTS

This work has been supported by FEDER and MEC, TEC2016-77791-C4-2-R, SIX (CZ.1.05/2.1.00/03.0072), and LOl401.

## 7. COMPLIANCE WITH ETHICAL STANDARDS

The authors declare that they have no conflict of interest.

All procedures performed in studies involving human participants were in accordance with the ethical standards of the institutional and/or national research committee and with the 1964 Helsinki declaration and its later amendments or comparable ethical standards. For this type of study formal consent is not required.

This chapter does not contain any studies with animals performed by any of the authors.

Informed consent was obtained from all individual participants included in the study.